\documentclass[10pt,twocolumn,letterpaper]{article}

\usepackage{cvpr}              %

\usepackage{amsmath}
\usepackage[T1]{fontenc}
\usepackage{siunitx}
\usepackage{booktabs}

\usepackage{graphicx}
\usepackage{multirow}

\definecolor{cvprblue}{rgb}{0.21,0.49,0.74}
\usepackage[pagebackref,breaklinks,colorlinks,allcolors=cvprblue]{hyperref}

\title{\raisebox{-0.5em}{\includegraphics[height=1.8em]{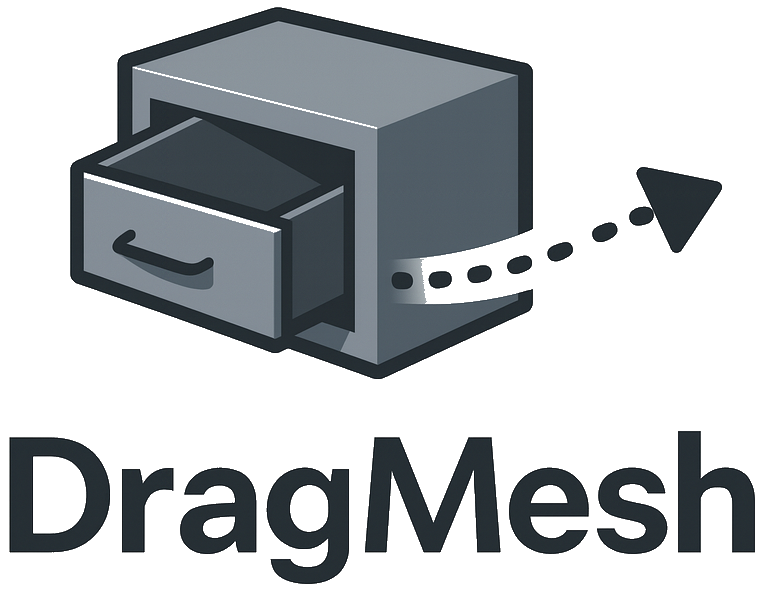}}~DragMesh: Interactive 3D Generation Made Easy}

\author{
    \textbf{Tianshan Zhang}$^{*}$ \quad
    \textbf{Zeyu Zhang}$^{*\dag}$ \quad
    \textbf{Hao Tang}$^{\ddag}$ \vspace{0.1cm}\\
    School of Computer Science, Peking University\\
    \small $^*$Equal contribution. $^\dag$Project lead.
    $^\ddag$Corresponding authors: bjdxtanghao@gmail.com.
}

\begin{document}

\twocolumn[{%
\renewcommand\twocolumn[1][]{#1}%
\maketitle

\includegraphics[width=\textwidth]{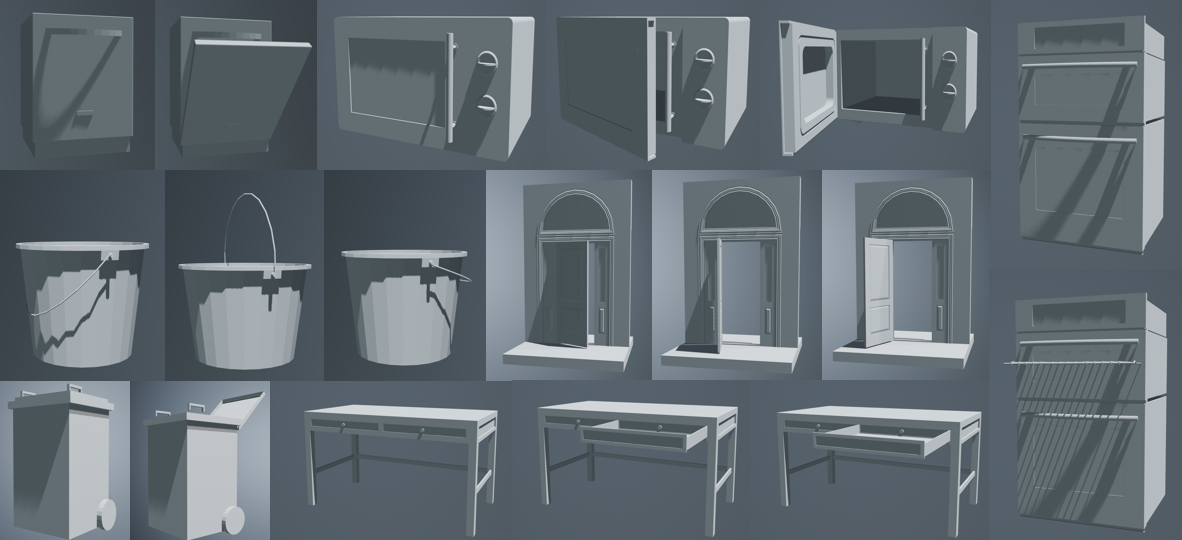}
\protect\captionof{figure}{\textbf{Results by DragMesh:} Our method translates intuitive drag-and-drop actions into accurate joint motion. It correctly infers and generates the motion of rotational joints (e.g., microwave, bucket, door) and translational joints (e.g., desk drawer, oven rack). For each object group, the images from left to right typically depict the initial state, an intermediate generated motion, and the final articulated state. 
Code: \url{https://github.com/AIGeeksGroup/DragMesh}
Website: \url{https://aigeeksgroup.github.io/DragMesh}}
\label{fig:teaser}
\vspace{3em}
}]

\begin{abstract}
   While generative models have excelled at creating static 3D content, the pursuit of systems that understand how objects move and respond to interactions remains a fundamental challenge. Current methods for articulated motion lie at a crossroads: they are either physically consistent but too slow for real-time use, or generative but violate basic kinematic constraints. We present DragMesh, a robust framework for real-time interactive 3D articulation built around a lightweight motion generation core. Our core contribution is a novel decoupled kinematic reasoning and motion generation framework. First, we infer the latent joint parameters by decoupling semantic intent reasoning (which determines the joint type) from geometric regression (which determines the axis and origin using our Kinematics Prediction Network (KPP-Net)). Second, to leverage the compact, continuous, and singularity-free properties of dual quaternions for representing rigid body motion, we develop a novel Dual Quaternion VAE (DQ-VAE). This DQ-VAE receives these predicted priors, along with the original user drag, to generate a complete, plausible motion trajectory. To ensure strict adherence to kinematics, we inject the joint priors at every layer of the DQ-VAE's non-autoregressive Transformer decoder using FiLM (Feature-wise Linear Modulation) conditioning. This persistent, multi-scale guidance is complemented by a numerically-stable cross-product loss to guarantee axis alignment. This decoupled design allows DragMesh to achieve real-time performance and enables plausible, generative articulation on novel objects without retraining, offering a practical step toward generative 3D intelligence.
\end{abstract}

\section{Introduction}

The pursuit of generative 3D intelligence—systems capable of autonomously constructing, understanding, and manipulating the physical world—has long been a central ambition of computer vision and graphics.
While 2D diffusion and large-scale vision-language models have made striking progress in image synthesis and scene understanding, extending this generative capability into the 3D domain remains fundamentally more challenging. Unlike pixels, 3D structures obey physical constraints, exhibit compositional part relationships, and evolve through spatially consistent motion. A complete 3D generative system must, therefore, not only represent what the world looks like but also how it moves and responds to interaction.

Recent advances such as implicit neural fields (NeRFs), 3D Gaussian Splatting, and mesh-based transformers have largely solved static reconstruction: we can now generate photorealistic geometry and texture at scale~\cite{mildenhall2020nerf,he2025sparseflex,kerbl20233d,liu2024meshformer}.
However, these methods remain fundamentally passive—they describe the world as it is, not how it behaves.
Static generative priors lack a notion of causal articulation or kinematic structure, making them unsuitable for real-world interactions, robotics, or design tasks that require physically grounded motion.

A natural next step for 3D generation is interactive articulation—learning how objects with movable parts respond to user or agent input.
This direction bridges static synthesis and embodied reasoning, enabling models that understand both form and function~\cite{chen2024partgen,geng2022gapartnet,guo2025articulatedgs,wu2025reartgs}. Yet, current solutions lie at two extremes: optimization-based methods that are physically consistent but too slow for real-time use, and large generative models that produce plausible motions but violate basic kinematic constraints.
The field thus faces a core dilemma between fidelity, physicality, and efficiency.

Pioneering works such as DragApart~\cite{li2024dragapart} initiated the study of interactive generation, demonstrating that user drags could drive object deformation~\cite{he2025sparseflex}.
However, they operated only on projected geometry, lacking true 3D structure and physical grounding.
Subsequent 3D Gaussian Splatting–based approaches like ArtGS~\cite{liu2025artgs} introduced explicit volumetric representations, achieving physically plausible effects but at the cost of scalability—each object requires a separate model.

Across these paradigms, a recurring limitation persists: interactive fidelity remains bound by computational cost and model specificity.
This dilemma arises because existing generalizable models attempt to solve two fundamentally different problems with one monolithic, heavyweight architecture: 1) Kinematic Reasoning (a one-time prediction of what can move) and 2) Motion Generation (a real-time synthesis of how it moves).

This motivates our decoupled, probabilistic formulation, built around a lightweight generation core that can generalize across diverse 3D meshes while enabling real-time interaction. Our paper presents three key contributions:

\begin{itemize}[itemsep=0pt, topsep=2pt]
    \item \textbf{Dual Quaternion VAE for Efficient Motion Learning.}  We introduce DualQuaternionVAE, a compact and efficient architecture that learns articulated motions via geometrically principled dual quaternion encoding by transforming only articulated parts with FiLM-based conditioning, our single-pass generation ensures both kinematic consistency and computational efficiency.
    
    \item \textbf{Decoupled Kinematic Reasoning and Motion Generation Framework.} We construct a decoupled framework that separates kinematic reasoning from generative motion synthesis. Our framework first separates semantic intent reasoning (determining the joint type) from geometric regression (which our Kinematics Prediction Network (KPP-Net) performs to find the axis and origin). The full predicted priors are then fed to the Dual-Quaternion VAE (DQ-VAE) to generate articulated motion.
    
    \item \textbf{Efficiency and Generalization Trade-off.} We compare our model with state-of-the-art methods, demonstrating that existing generalizable methods are computationally expensive, requiring 5-10 times more parameters and GFlops of computation. Conversely, other lightweight methods lack generalization ability and require individual training for each object. Our contribution is that DragMesh implements an optimally balanced framework, whose core generation module achieves robust generalization to new objects with low computational overhead.
    
\end{itemize}

In summary, DragMesh bridges the gap between large-scale generative modeling and real-time interactive articulation, offering a practical step toward generative 3D intelligence.

\section{Related Work}
\subsection{3D Generation}

The evolution of 3D generation has been driven by a persistent tension between representation fidelity and downstream editability. Early voxel-based methods~\cite{Peng_2022_CVPR, chen2023shaddr, Zhou_2021_ICCV} and implicit representations like DeepSDF~\cite{Park_2019_CVPR} and NeRF~\cite{mildenhall2020nerf} prioritized visual quality, achieving photorealistic reconstruction through dense volumetric encoding. However, their monolithic, continuous formulations produced "black-box" outputs that were inherently resistant to structural editing or part-level manipulation. Transformer-based reconstruction models (LRM~\cite{hong2023lrm}, TripoSR~\cite{tochilkin2024triposr}) and diffusion pipelines (TripoSG~\cite{li2025triposg}) have since scaled these approaches to unprecedented fidelity; yet, they perpetuate the same fundamental limitation: they generate holistic, unstructured geometries that lack explicit part decomposition.

To unlock editability, recent work has revisited explicit mesh representations~\cite{siddiqui2024meshgpt, Lionar_2025_CVPR, li2024craftsman3d, xu2024instantmesh, wang2024llamameshunifying3dmesh, zhao2025deepmesh, chen2024meshanything, chen2024meshanything2}. Unlike neural fields, meshes expose vertex-face topology directly, enabling fine-grained vertex editing, skeletal rigging, and seamless integration with physics engines and rendering pipelines. This explicit structure is not merely a convenience—it is essential for downstream applications in animation, simulation, and interactive design.

Our DragMesh builds directly upon these structured mesh and part-level generative foundations by taking their decomposable meshes as inputs and enabling real-time, physically consistent manipulation through a lightweight dual-quaternion generative model.

\subsection{Interactive Manipulation}

The paradigm of point-based interactive editing, pioneered in 2D by methods~\cite{pan2023_DragGAN, shi2024dragdiffusion, wu2024draganything,zhang2023adding, peng2024controlnext}, cannot be directly applied to 3D scenarios that demand multi-view consistency and adherence to physical constraints, as these methods fundamentally manipulate pixels rather than underlying geometry. To address this, a common paradigm is the "2D-3D-2D" framework, where a 2D object is first "lifted" to a 3D representation, edited in a 3D space where physical constraints can be applied, and then reprojected back into the original~\cite{chen2024mvdrag3d, xie20252d, pandey2024diffusion}. The existence of this complex workflow demonstrates the inherent limitations of direct 2D manipulation for 3D-consistent tasks.

Manipulating objects with movable parts, such as opening a drawer or folding a laptop, presents a level of complexity far beyond the scope of standard 3D generation models. To address this, some approaches have proposed using Transformer-based methods to autoregressively generate interactive components. Methods like MeshArt~\cite{gao2025meshart}, ArtFormer~\cite{su2025artformer}, FreeArt3D~\cite{chen2025freeart3d}, Articulate-Anything~\cite{qiu2025articulate}, and DIPO~\cite{wu2025dipo} focus specifically on generating objects with explicit parts and joints. DragAPart~\cite{li2024dragapart} learns part-level motion priors to predict plausible deformations from user drags. Even advanced representations like 3D Gaussian Splatting (3DGS) require specialized extensions, such as ArtGS~\cite{liu2025building}, Part2GS~\cite{yu2025part} and RoboSimGS~\cite{zhao2025high}, which integrate physical modeling and part-aware representations to impose constraints and handle articulated motion. However, diffusion-based models, while capable of generating high-fidelity 3D results, create an "Interaction-Fidelity Chasm." Their iterative sampling process, which can take several minutes, forces users to choose between near-instant feedback with low-quality, non-generative methods (like direct mesh deformation) or high-fidelity generative results at the cost of significant latency~\cite{geng2022gapartnet, gao2025partrm}. This latency significantly disrupts the seamless, iterative creative workflow that is standard in 2D design software, where low-latency, real-time adjustments are expected.

\begin{figure}[t]
    \centering
    \includegraphics[width=\linewidth]{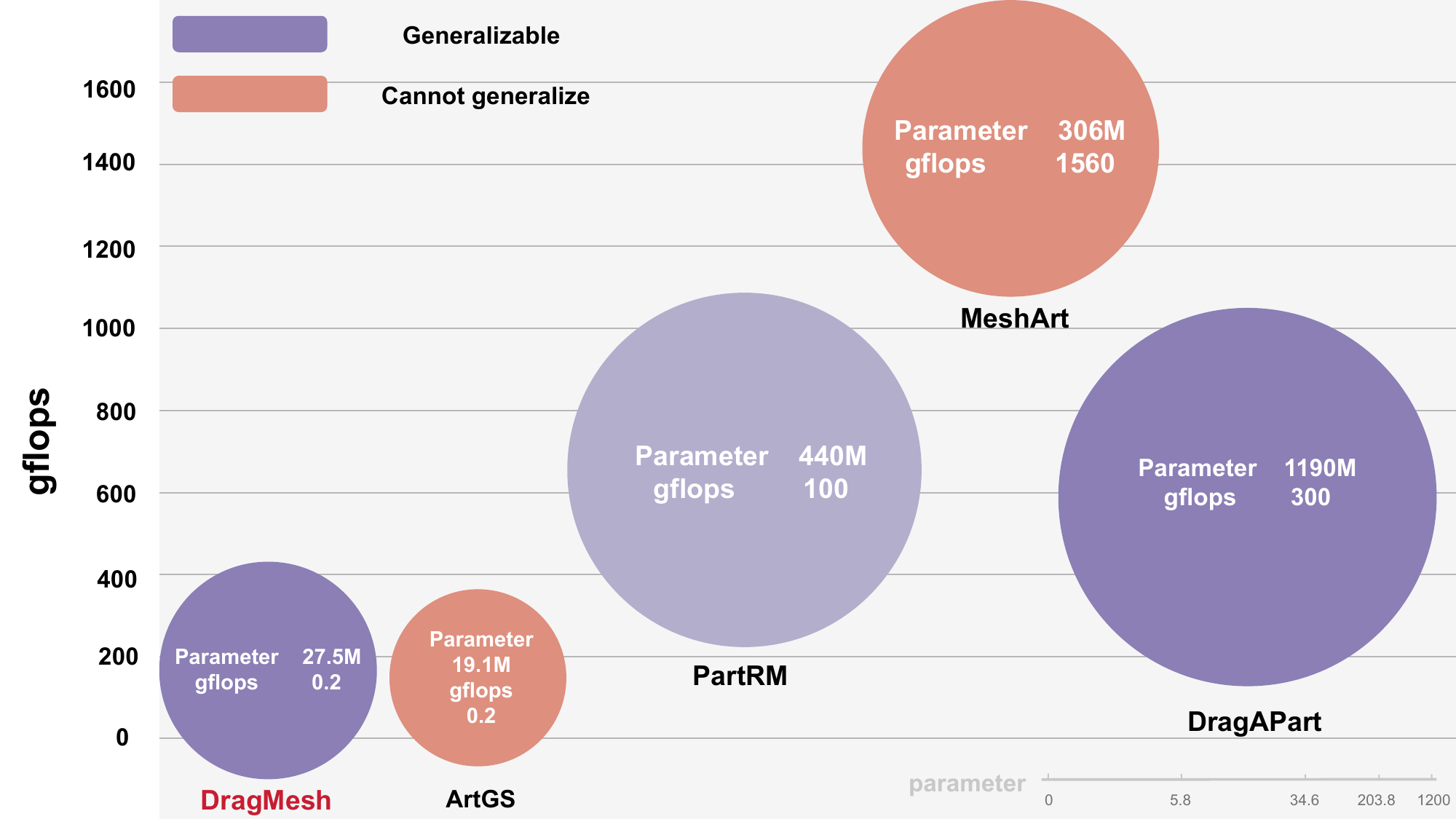} 
    \caption{Bubble size indicates parameter count (core module only), color denotes generalization capability (Purple: generalizable; Orange: per-object training required). Our method achieves generalization with significantly lower computational cost than existing generalizable approaches.}
    \label{fig:model_comparison}
\end{figure}

\section{Methodology}
\label{sec:method}

\subsection{Overview}

\begin{figure*}[t]
    \centering
    \includegraphics[width=\linewidth]{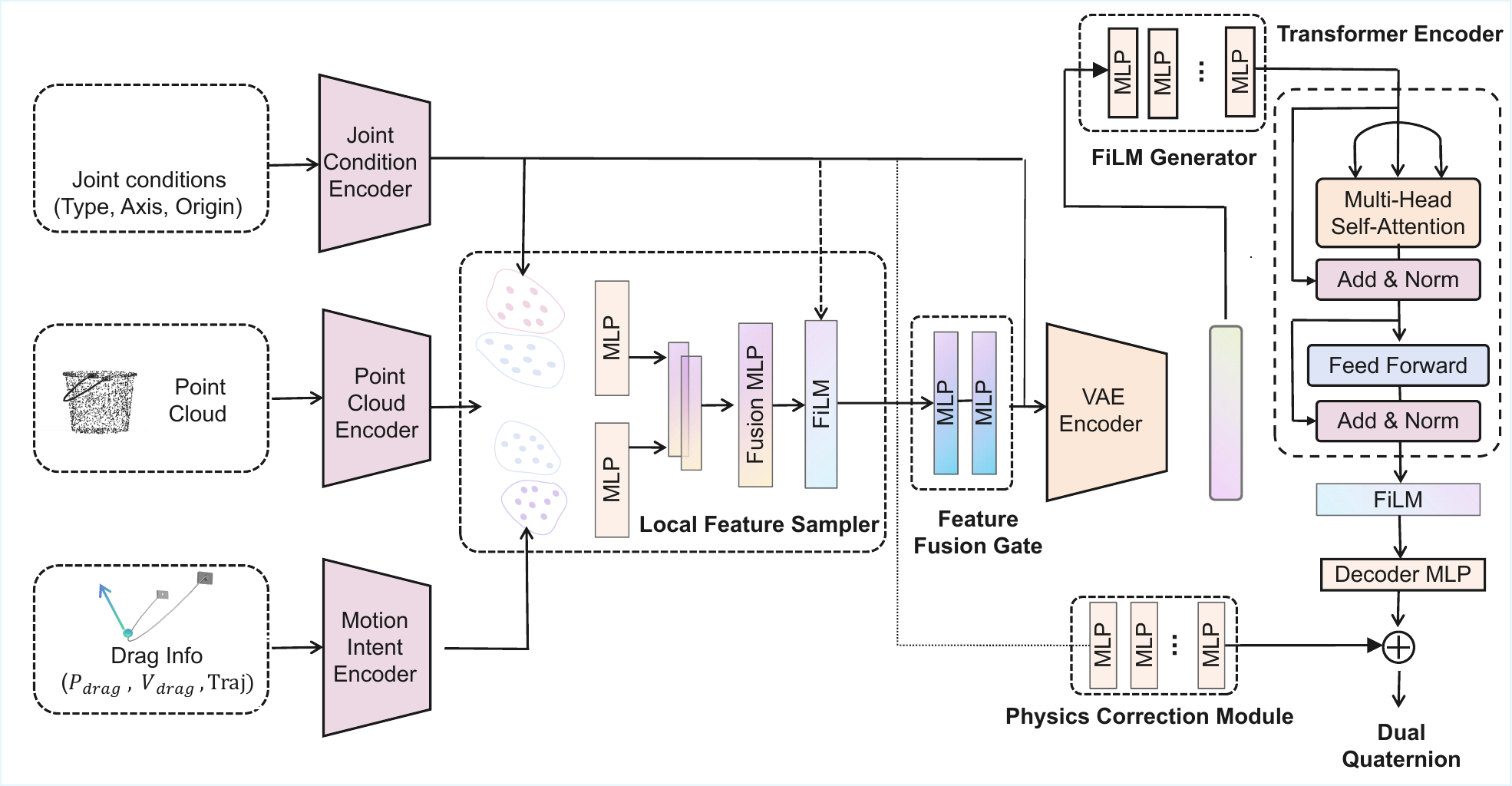} 
    \caption{The DragMesh pipeline fuses point cloud, joint, and drag inputs through a VAE and Transformer architecture to predict a physically-corrected dual quaternion.
    }
    \label{fig:pipeline}
\end{figure*}

Our training framework (illustrated in Figure~\ref{fig:pipeline}) decouples motion generation from kinematic reasoning. This design provides kinematic consistency by using dual quaternions and enables efficient, real-time inference by transforming only articulated parts.

A Dual Quaternion VAE (DQ-VAE): A conditional generative model (Sec.~\ref{model}) that outputs physically consistent motion trajectories. It is trained to produce dual quaternion sequences representing the articulation.

A Kinematics Prediction Pipeline: To enable annotation-free use, this pipeline first uses a VLM to determine the semantic interaction type. Then, our proposed Kinematics Prediction Network (KPP-Net, Sec~\ref{KPP-Net}) regresses the precise geometric parameters.

At inference, the parameters predicted by this pipeline are fed into the pre-trained DQ-VAE, enabling generative articulation on novel meshes.

\subsection{Motion Representation and Pre-processing}

Traditional representations suffer from inherent drawbacks: Euler angles exhibit gimbal lock, axis-angle lacks translation encoding, and transformation matrices require 12 parameters per frame. Dual quaternions (DQ) provide a compact (8-parameter), continuous, and singularity-free parameterization for rigid body transformations, naturally satisfying screw motion theory.

Formally, given a joint with unit axis $\mathbf{a} \in \mathbb{R}^3$ and origin $\mathbf{o} \in \mathbb{R}^3$, we generate ground-truth motion sequences of length $T=16$ by uniformly sampling angles $\theta_t \in [0, \theta_{\max}]$ (for revolute joints) or distances $d_t$ (for prismatic joints), where $\otimes$ denotes quaternion multiplication, and the dual quaternion at time $t$ is:
\begin{equation}
\begin{array}{l}
    \mathbf{q}_r^{(t)} = [\cos(\theta_t/2), \mathbf{a}\sin(\theta_t/2)], \\[4pt]
    \hspace{1em}\mathbf{q}_d^{(t)} = \tfrac{1}{2}\mathbf{q}_r^{(t)} \otimes [0, \mathbf{t}_{\text{trans}}]
\end{array}
\end{equation}
where $\mathbf{t}_{\text{trans}} = -\mathbf{o} + \mathbf{q}_r^{(t)} \otimes \mathbf{o}$ 
encodes the translation induced by rotating around point $\mathbf{o}$.

For a prismatic joint:
\begin{equation}
    \mathbf{q}_r^{(t)} = [1, 0, 0, 0], \quad \mathbf{q}_d^{(t)} = \frac{1}{2}[0, d_t\mathbf{a}]
\end{equation}

To establish a canonical input space, we normalize each mesh $\mathcal{M}$ by its axis-aligned bounding box center $\mathbf{c}$ and largest dimension $s$. All spatial quantities (vertices, drag point $\mathbf{p}_{\text{drag}}$, joint origin $\mathbf{o}$) are transformed via $(\cdot - \mathbf{c})/s$, while vectors (drag vector $\mathbf{v}_{\text{drag}}$, the translational component of $\mathbf{q}_d^{\text{gt}}$) are scaled by $1/s$. The joint axis $\mathbf{a}$ remains a unit vector.

\subsection{Network Architecture}
\label{model}
In this section, we formalize our probabilistic articulation framework.
Given an input mesh and user drag signal, our goal is to learn a distribution 
$p(\mathbf{q}_r, \mathbf{q}_d | \mathbf{x}, \mathbf{d})$ over dual quaternions representing rigid-body transformations,
where $\mathbf{x}$ denotes geometric features and $\mathbf{d}$ denotes interaction cues.
We achieve this through a variational autoencoder that jointly encodes geometric and kinematic priors.

\subsubsection{Multi-Modal Encoders}
We process the three input modalities using specialized encoders:
\begin{enumerate} 
    \item \textbf{Joint Condition Encoder.} This module encodes the fundamental kinematic constraints. It processes the discrete Joint Type (revolute/prismatic) through an Embedding layer, and the continuous Joint Axis $\mathbf{a}$ and Joint Origin $\mathbf{o}$ through separate MLPs. These are fused into a 512-dimensional joint feature vector, $\mathbf{f}_{\text{joint}}$.
    \item \textbf{Point Cloud Encoder.} We sample 4096 points from the normalized mesh, concatenating the 3D coordinates (XYZ) with a binary Part Mask (indicating which part moves) to form a 4D input. A PointNet-style encoder (PointCloudEncoder) processes this 4D point cloud to extract per-point features $\mathbf{f}_{\text{points}} \in \mathbb{R}^{N \times 1024}$.
    \item \textbf{Motion Intent Encoder.} This module interprets the user's drag interaction. It dynamically computes a feature $\mathbf{f}_{\text{motion}}$ based on the joint type. For revolute joints, it combines features from the Rotation Direction and the Average Trajectory Velocity (processed by separate MLPs). For prismatic joints, it uses features derived from the Drag Chord using a MultiScaleDragEncoder. An Amplitude feature is added to all types. Finally, a separate Drag Trajectory feature, encoded via a small Transformer, is concatenated with $\mathbf{f}_{\text{motion}}$ to capture the path's temporal shape regardless of joint type.
\end{enumerate}

\subsubsection{Feature Fusion and Conditional VAE}

Instead of simple concatenation, we fuse features in a structured manner to preserve local and global information. To capture geometry relevant to the interaction, we sample local features from $\mathbf{f}_{\text{points}}$ using k-Nearest Neighbors (kNN) around the joint origin and drag point. The outputs are processed by MLPs and fused. Critically, this module is modulated by $\mathbf{f}_{\text{joint}}$ via a FiLM layer, allowing the joint type to influence which local features are salient. This produces a context-aware local feature $\mathbf{f}_{\text{local}}$.

A learned gating mechanism combines the outputs from all preceding modules. This MLP takes the concatenation of $[\mathbf{f}_{\text{local}}, \mathbf{f}_{\text{joint}}, \mathbf{f}_{\text{motion}}]$ and produces a single fused motion-context feature, $\mathbf{f}_{\text{fused}}$.

To parameterize the latent space, the VAE encoder receives a carefully constructed input: the concatenation of the fused feature and the raw joint feature, $[\mathbf{f}_{\text{fused}}, \mathbf{f}_{\text{joint}}]$. This structure ensures the VAE is always strongly conditioned on the primary joint constraints, even after fusion. The encoder then outputs the latent parameters $\boldsymbol{\mu}$ and $\log\boldsymbol{\sigma}^2$.

\begin{equation}
\begin{array}{l}
    \mathbf{f}_{\text{combined}} = \text{Concat}(\mathbf{f}_{\text{fused}}, \mathbf{f}_{\text{joint}}), \\[4pt]
    \boldsymbol{\mu}, \log \boldsymbol{\sigma}^2 = \text{MLP}_{\text{enc}}(\mathbf{f}_{\text{combined}}), \\[4pt]
    \mathbf{z} = \boldsymbol{\mu} + \boldsymbol{\sigma} \odot \boldsymbol{\epsilon}, \quad 
    \boldsymbol{\epsilon} \sim \mathcal{N}(0, \mathbf{I})
\end{array}
\end{equation}

\subsubsection{Conditioned Transformer Decoder and Physics Correction}

Our decoder must translate the latent code $\mathbf{z}$ into a temporally coherent motion sequence $\mathbf{Q} \in \mathbb{R}^{T \times 8}$ while adhering to joint constraints. A naive MLP decoder is insufficient, as it fails to model inter-frame correlations, leading to unrealistic, drifting motion. We also found standard autoregressive models unsuitable, as they suffer from error accumulation, where early errors cascade, and latent space collapse, where the decoder learns to ignore $\mathbf{z}$, thus hindering the VAE's representation learning.

We therefore employ a non-autoregressive Transformer Encoder, which processes all $T=16$ frames in parallel. This design captures global sequence coherence without the fragility of autoregressive models and ensures every frame is robustly conditioned on $\mathbf{z}$. To strongly enforce the kinematics, the decoder input sequence is formed by concatenating the positional encoding, the repeated latent code $\mathbf{z}$, and multiple, scaled injections of the joint feature $\mathbf{f}_{\text{joint}}$. This strong conditioning is maintained throughout the network's depth using FiLM (Feature-wise Linear Modulation), where $\mathbf{f}_{\text{joint}}$ predicts affine parameters $(\boldsymbol{\gamma}, \boldsymbol{\beta})$ to modulate each Transformer layer's output.

The Transformer's output is first passed to an MLP to predict a base sequence, $\mathbf{Q}_{\text{base}}$. This is then refined by a Physics Correction module, which predicts a residual $\mathbf{Q}_{\text{residual}}$ such that the final output is $\mathbf{Q}_{\text{final}} = \mathbf{Q}_{\text{base}} + \mathbf{Q}_{\text{residual}}$. To provide maximum context for this correction, its input includes $\mathbf{Q}_{\text{base}}$, the joint type embedding, and both raw and scaled versions of the joint feature $\mathbf{f}_{\text{joint}}$ and the joint axis $\mathbf{a}$, ensuring the final motion strictly respects the physical constraints.

\subsection{VAE Training Objectives}
\label{train}
Our objective is to train the VAE to generate motions that are not only geometrically accurate but also physically plausible and consistent with the specified joint constraints.

We found that a naive baseline objective, consisting only of a geometric loss ($\mathcal{L}_{\text{cd}}$) and a direct quaternion reconstruction loss ($\mathcal{L}_{\text{qr}}, \mathcal{L}_{\text{qd}}$), was insufficient. With only these losses, the VAE struggles to learn a meaningful and well-structured latent representation. The network fails to respect the underlying kinematics, often producing implausible motions, such as drifting or arbitrary rotations, rather than the intended motion around the specified joint axis. We found it essential to introduce a set of explicit physical constraint losses. 

To resolve this, we found it essential to introduce a set of explicit physical constraint losses. These constraints serve as a strong inductive bias, forcing the network to learn the correct motion laws. Our final loss $\mathcal{L}$ is a weighted sum of four key components:
\begin{equation}
    \begin{aligned}
        \mathcal{L} = & \;\; \lambda_{\text{mesh}} \mathcal{L}_{\text{mesh}} + \lambda_{\text{quat}} (\mathcal{L}_{\text{qr}} + \mathcal{L}_{\text{qd}}) + \lambda_{\text{cd}} \mathcal{L}_{\text{cd}} \\
        & + \lambda_{\text{axis}} \mathcal{L}_{\text{axis}} + \lambda_{\text{qd0}} \mathcal{L}_{\text{qd0}} + \lambda_{\text{qr1}} \mathcal{L}_{\text{qr1}}  + \lambda_{\text{kl}} \mathcal{L}_{\text{KL}}
    \end{aligned}
\end{equation}

\paragraph{Reconstruction and Geometric Losses.} We supervise the output at two complementary levels. $\mathcal{L}_{\text{mesh}}$ (L2 vertex loss) and $\mathcal{L}_{\text{cd}}$ (Chamfer Distance) ensure fine-grained geometric accuracy for the moving part. Concurrently, $\mathcal{L}_{\text{qr}}$ and $\mathcal{L}_{\text{qd}}$ provide direct supervision on the motion parameters themselves, using an L2 loss on the extracted 3D translation vector $\mathbf{t}$ and a dot-product-based distance for rotation. For all reconstruction losses, we apply a 2x weight to the first frame ($t=0$) to enforce an accurate starting pose.

\paragraph{Physical Constraint Losses.} This is the core component that enforces kinematic plausibility. $\mathcal{L}_{\text{axis}}$ dynamically enforces the joint's degrees of freedom: for revolute joints, it penalizes misalignment between the predicted rotation axis and the ground-truth axis $\mathbf{a}$; for prismatic joints, it penalizes any translation component perpendicular to $\mathbf{a}$. Furthermore, we add two zero-motion penalties: $\mathcal{L}_{\text{qd0}}$ penalizes any translation for revolute joints, while $\mathcal{L}_{\text{qr1}}$ penalizes any rotation for prismatic joints.

\paragraph{VAE Regularization with Free Bits.}
Finally, $\mathcal{L}_{\text{KL}}$ regularizes the latent space. To prevent the "posterior collapse" common in VAEs (where the decoder ignores the latent code), we employ a Free Bits strategy. We define a minimum information threshold $\delta$ and only penalize the KL divergence for samples that fall below this threshold:
\begin{equation}
\mathcal{L}_{\text{KL}} = \frac{1}{B}\sum_{b} \max(0, \text{D}_{\text{KL}}(q(\mathbf{z}|\cdot) | p(\mathbf{z})) - \delta)
\end{equation}
This encourages the VAE to encode meaningful information before being forced to match the prior, which resolves the training difficulties we initially observed.

\subsection{Kinematics Prediction Network (KPP-Net)}
\label{KPP-Net}
To enable annotation-free articulation at inference, we train a separate network, KPP-Net, to predict kinematic parameters from geometry and user interaction. Unlike the Dual-Quaternion VAE, which models motion generation, KPP-Net focuses on estimating the underlying joint configuration—specifically the axis and origin—from an articulated mesh, its part segmentation, and a drag interaction.

KPP-Net follows a point-based design, composed of a Transformer-based Point Encoder. This architecture uses two parallel encoders to extract features: a Global Encoder processes a 10D input (XYZ + part mask + repeated drag point/vector), while a Local Encoder processes the part-masked mesh to focus on local geometry. The resulting global and local features are concatenated and fed into two independent regression heads, which predict the joint axis and origin respectively:
\begin{equation}
    [\hat{\mathbf{a}}, \hat{\mathbf{o}}] = f_{\text{KPP}}(\mathbf{P}, \mathbf{M}, \mathbf{d}_p, \mathbf{d}_v)
\end{equation}
where $\mathbf{P}\!\in\!\mathbb{R}^{N\times3}$ denotes sampled points, $\mathbf{M}$ the binary part mask, and $\mathbf{d}_p, \mathbf{d}_v$ the drag interaction vectors.

KPP-Net is trained independently using ground-truth URDF parameters with a composite loss function combining weighted axis and origin losses:
\begin{equation}
    \mathcal{L}_{\text{KPP}} = \lambda_{\text{axis}} \mathcal{L}_{\text{axis}} + \lambda_{\text{origin}} \mathcal{L}_{\text{origin}}
\end{equation}
$\mathcal{L}_{\text{axis}}$ is the geodesic loss, $\arccos(\text{clamp}(|\hat{\mathbf{a}} \cdot \mathbf{a}_{gt}|, 0.0, 1.0 - \epsilon))$, which measures the angle between the predicted and ground-truth axes. $\mathcal{L}_{\text{origin}}$ is the Smooth L1 Loss, which robustly regresses the origin coordinates. This formulation encourages the predicted joint axis to align with the true kinematic direction while accurately localizing the joint origin.

\subsection{Inference Pipeline}
\label{inference pipeline}
During inference, DragMesh operates in a fully annotation-free setting, taking as input only a raw mesh $\mathcal{M}$ and a user-defined drag interaction $(\mathbf{p}_{\text{drag}}, \mathbf{v}_{\text{drag}})$. The goal is to generate a physically consistent articulated motion without any URDF annotations. The complete process proceeds in three stages.

\begin{figure*}[t]
    \centering
    \includegraphics[width=\linewidth]{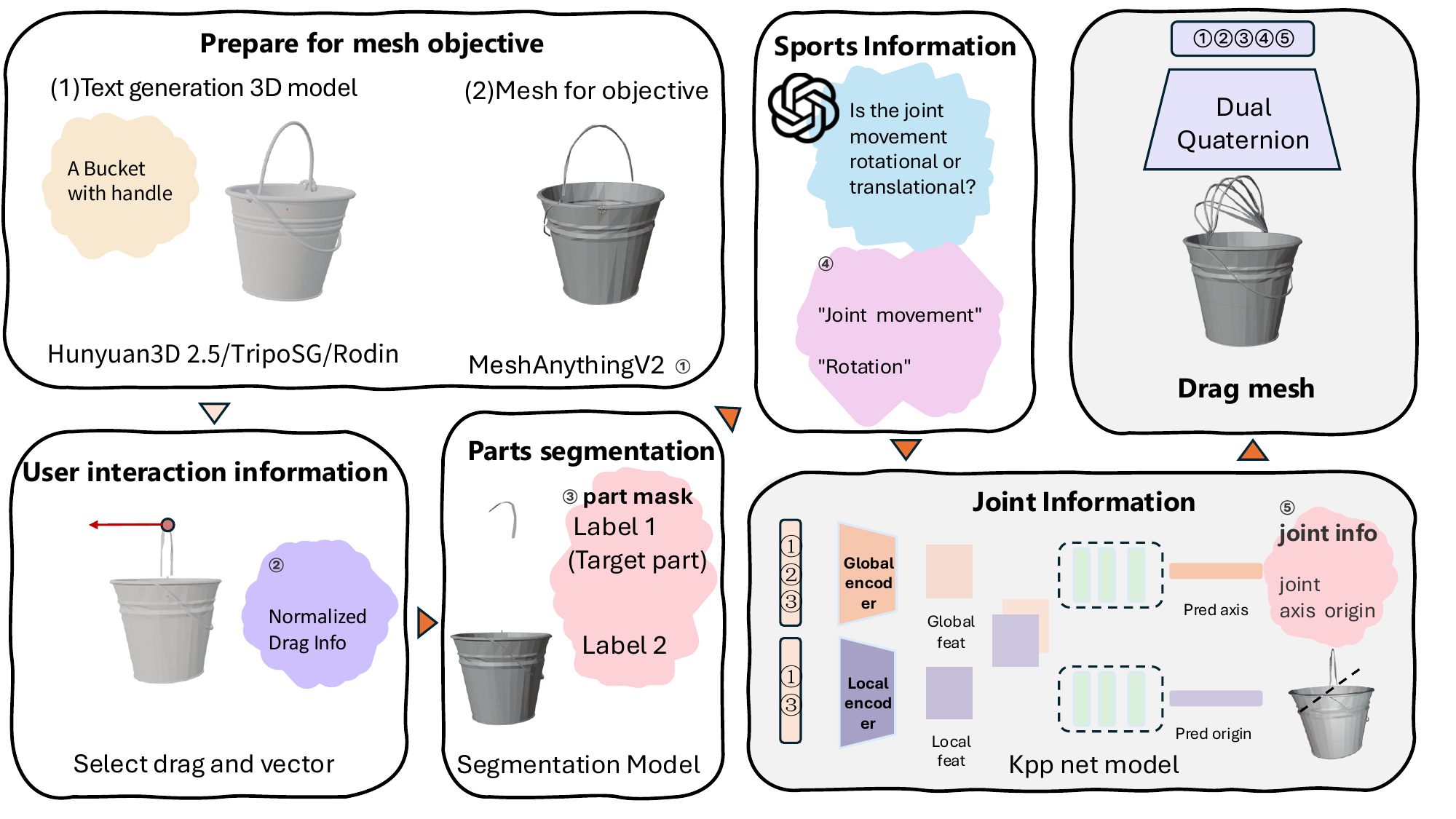}
    \caption{The DragMesh Annotation-Free Inference Pipeline. Given a raw mesh and drag, a segmentation model identifies the movable part while a VLM predicts the joint type. Our KPP-Net then regresses the precise axis and origin, enabling the final DragMesh model to generate the Dual Quaternion animation.}
    \label{fig:inference_pipeline}
\end{figure*}

\paragraph{Part Segmentation.}
We segment the raw mesh $\mathcal{M}$ into movable components using an off-the-shelf part segmentation model (e.g., P3-SAM~\cite{ma2025p3}). This yields a discrete part mask $\mathbf{M}$. The user’s drag point $\mathbf{p}_{\text{drag}}$ is used to identify the \textit{target part} $\mathcal{M}_{\text{mov}}$ (with mask $\mathbf{M}=1$) from the static subset $\mathcal{M}_{\text{sta}}$ (with mask $\mathbf{M}=0$). This step provides the structural decomposition required for kinematic reasoning.

\paragraph{Intent Reasoning and Kinematic Prediction.}We found that while our KPP-Net (Sec.~\ref{KPP-Net}) excels at geometric regression, it struggles to reliably predict the motion intent (\textit{i.e.,} revolute vs. prismatic) when the input only contains mesh information. Relying solely on global point cloud data is insufficient to resolve this inherent semantic ambiguity.To address this, we decouple the task. First, we use a VLM (e.g., GPT-4o) to provide the necessary semantic context by processing information about the interacting parts. The VLM robustly determines the high-level interaction intent, outputting the categorical joint type $j_{\text{type}}$. This predicted $j_{\text{type}}$ is then fed into our pre-trained KPP-Net. We leverage KPP-Net's strength in geometric regression, using only its axis and origin heads to predict the precise kinematic parameters $\hat{\mathbf{a}}$ and $\hat{\mathbf{o}}$. This combined approach provides the full kinematic constraints $\hat{\mathbf{c}} = (j_{\text{type}}, \hat{\mathbf{a}}, \hat{\mathbf{o}})$ required by the VAE.

\paragraph{Articulated Motion Generation.}Finally, the full set of predicted kinematic parameters $\hat{\mathbf{c}}$, along with the user drag features $\mathbf{f}_{\text{drag}}$ (e.g., $\mathbf{d}_v$ and an interpolated trajectory) and mesh features, are fed into the trained Dual Quaternion VAE. The decoder produces a temporal sequence of dual quaternions $\{(\mathbf{q}_r^{(t)}, \mathbf{q}_d^{(t)})\}_{t=1}^{T}$.To realize the animation, we apply these transformations selectively to the movable vertices $\mathcal{M}_{\text{mov}}$:
\begin{equation}
\mathbf{v}_i^{(t)} =
\begin{cases}
    \text{DQApply}\big((\mathbf{q}_r^{(t)}, \mathbf{q}_d^{(t)}), \mathbf{v}_i, \hat{\mathbf{o}}\big), & \text{if } \mathbf{M}_i = 1, \\
    \mathbf{v}_i, & \text{if } \mathbf{M}_i = 0.
\end{cases}
\end{equation}

where $\text{DQApply}$ denotes the rigid transformation applied relative to the predicted origin $\hat{\mathbf{o}}$. This completes the pipeline, enabling DragMesh to animate unseen meshes from raw inputs by combining LLM-based intent reasoning with KPP-Net's geometric prediction.

\section{Experiments}
\subsection{Datasets and Evaluation Metrics}

\paragraph{Datasets.}
There exist multiple datasets containing articulated objects and joint annotation~\cite{Xiang_2020_SAPIEN, Mo_2019_CVPR,liu2022akb48realworldarticulatedobject,gao2025partrm}, reflecting the growing interest in modeling 3D object articulation. We chose GAPartNet~\cite{geng2022gapartnet}, which provides high-quality interaction models with part-level geometry, joint structures defined by URDF, and physically valid mobility parameters. It covers a wide range of real-world categories and provides rich metadata and diverse articulated models.

\paragraph{Metrics.}

We evaluate: (1) \textit{Geometric accuracy} via Chamfer Distance (CD)~\cite{fan2017point} and Vertex L2 Error ($\mathcal{L}_{mesh}$); (2) \textit{Physical constraints} via Axis Error (milliradians) and Origin Error (millimeters) for joint parameters, plus zero-motion penalties for invalid revolute translations and prismatic rotations; (3) \textit{Latent quality} via KL divergence to detect posterior collapse.

\subsection{Comparative Study}
\label{Comparative Study}
Our goal is to provide a horizontal comparison of various interactive 3D content generation methods ~\cite{liu2025artgs, gao2025partrm,li2024dragapart}. We uniformly adopt the Objaverse dataset as our benchmark.

We observe that existing methods exhibit significant differences in their data requirements and preprocessing pipelines. For instance, some methods may require specific part annotations, watertight meshes, or are restricted to certain object categories. Therefore, forcing a single, fixed data subset for training all models would fail to fairly evaluate the optimal performance of each method. Consequently, our adopted strategy is as follows: We adhere to the official implementation of each competing method, preparing and filtering a specific data subset for each from our pre-screened Objaverse interactive corpus. We will detail the specific data processing steps taken for each method in the appendix. Given the differences in input data and training setups, our comparison will primarily focus on three aspects: the complexity of data processing, computational overhead (performance consumption), and the qualitative fidelity of the final generated results. Detailed data is available in the supplementary material~\ref{Comparative Study Detail}.

\subsection{Ablation Study}

We validate our design choices through comprehensive ablations on model architecture (Table~\ref{tab:additive_ablation_final}), loss functions (Table~\ref{tab:loss_ablation}), and the KPP geometry predictor (Table~\ref{tab:kpp_run_ablation}).

\begin{table*}[t] \small
\centering
\caption{\textbf{Additive Ablation of Model Architecture.} Each row progressively adds one component to the previous configuration.}
\label{tab:additive_ablation_final}
\resizebox{\textwidth}{!}{%
\begin{tabular}{l l S[table-format=3.3] S[table-format=2.5] S[table-format=1.4] S[table-format=2.4]}
\toprule
& & \multicolumn{1}{c}{\textbf{Reconstruction Error} $\downarrow$} & \multicolumn{2}{c}{\textbf{Physical Constraint Error} $\downarrow$} & \multicolumn{1}{c}{\textbf{VAE}} \\

\cmidrule(lr){3-3} \cmidrule(lr){4-5} \cmidrule(lr){6-6}
\textbf{ID} & \textbf{Model Variant (Progressive Build)} & {CD ($\times 10^{-3}$) $\downarrow$} & {Axis Error ($\times 10^{-3}$) $\downarrow$ } & {QR1 (Pris) $\downarrow$} & {KL (Raw)} \\
\midrule

M1 & \textbf{Base:} VAE + MLP Decoder & {...} & {...} & {...} & {2.68} \\
M2 & M1 + Transformer Decoder (Base) & {93.5} & {0.28} & {0} & {2.578} \\
M3 & M2 + Encoder Fusion (LFS + FFG) & {184.4} & {0.12} & {0.0017} & {18.88} \\
M4 & M3 + LFS FiLM Encoder & {78.773} & {16.74} & {1.457} & {29.85} \\
M5 & M4 + Transformer FiLM & {63.561} & {0.45593} & {0.0004} & {36.96} \\
M6 & M5 + Physics Correction (PCM) & {71.814} & {0.344} & {0} & {64.55} \\
\midrule
M7 & M6 + Physical Loss Terms   ~\textbf{(Ours)}  & \textbf{61.485} & \textbf{0.265} & \textbf{0} & \textbf{26.0515} \\
\bottomrule
\end{tabular}
}
\vspace{-0.2cm}
\end{table*}

\paragraph{Architecture Components.}
Table~\ref{tab:additive_ablation_final} reveals a critical failure mode in the baseline (M2): despite a low physics error, the KL value indicates that the model did not learn the correct information, generating only minimal motion (<5°).
Adding encoder fusion and FiLM conditioning (M3-M5) enables expressive motion but initially destabilizes physical accuracy. Our full model (M7) resolves this tension through Physics Correction (M6) and tailored loss terms, achieving optimal reconstruction, physical plausibility, and motion expressiveness.

\begin{table}[t] \small
\centering
\caption{\textbf{Loss Function Ablation.} Baseline uses only reconstruction and geometry losses with standard KL.}
\label{tab:loss_ablation}
\small
\begin{tabular}{l c c c c}
\toprule
\textbf{Model} & 
\multicolumn{2}{c}{\textbf{Physical Error} $\downarrow$} & 
\multicolumn{2}{c}{\textbf{Recon. \& VAE}} \\
\cmidrule(r){2-3} \cmidrule(l){4-5}
& Axis $\downarrow$ & Rev. Trans. $\downarrow$ & CD $\downarrow$ & KL \\
\midrule
(1) Baseline & 1.301 & 0.130 & 0.0577 & 67.35 \\
(2) + Physics & 0.0006 & \textbf{0.014} & 0.0653 & 34.56 \\
(3) + Free Bits & 0.680 & 0.080 & 0.0679 & 21.49 \\
(4) \textbf{Ours (Full)} & \textbf{0.0004} & \textbf{0.014} & \textbf{0.051} & \textbf{12.90} \\
\bottomrule
\end{tabular}
\end{table}

\paragraph{Loss Components.}
Table~\ref{tab:loss_ablation} demonstrates clear synergy between our physics losses and free-bits KL.
Physics losses alone (2) drastically reduce constraint violations but degrade reconstruction.
Free bits alone (3) improve VAE stability but fail to enforce physical correctness.
Combining both (4) yields the best results across all metrics, indicating that the free bits provide the capacity for complex motions while physics losses guide geometrically accurate and physically plausible solutions.

\begin{table}[t] \small
\centering
\caption{\textbf{KPP Geometry Predictor Ablation.} Progressive improvements.}
\label{tab:kpp_run_ablation}
\begin{tabular}{l c c}
\toprule
\textbf{Configuration} & \textbf{Axis (mrad)} $\downarrow$ & \textbf{Origin (mm)} $\downarrow$ \\
\midrule
Baseline (PointNet) & 450.0 & 15.0 \\
+ Mask & 350.0 & 12.0 \\
+ Drag & $\sim$300 & $>$10.0 \\
+ Attention Encoder & 150.0 & 6.0 \\
+ Decoupled Heads & 80.0 & 3.5 \\
\textbf{Ours (Full)} & \textbf{45.0} & \textbf{1.8} \\
\bottomrule
\end{tabular}
\end{table}

\paragraph{KPP Geometry Predictor.}
Table~\ref{tab:kpp_run_ablation} reveals that architectural improvements dominate performance gains.
Starting from a PointNet baseline with only mesh geometry, progressively adding mask and drag features (rows 1-3) yields modest improvements, with the drag feature even degrading origin prediction.
The critical breakthroughs come from architectural changes: replacing PointNet with a dual-stream attention encoder (row 4) achieves 2$\times$ error reduction, and decoupling the prediction heads (row 5) provides another 50\% improvement.
Note that rows 4-6 all use the complete feature set (mesh, mask, and drag), isolating the impact of architectural design on kinematic regression accuracy.

\section{Efficiency}

Figure~\ref{fig:model_comparison} illustrates the trade-offs between our method and state-of-the-art (SOTA) methods in terms of parameter count and GFLOPs in the core generation module. Clearly, existing generalizable models (purple bubbles) incur 5 to 10 times higher computational costs than ours due to their attempt to solve all problems with a single, large end-to-end model. Other lightweight methods (orange bubbles) sacrifice generalization ability, requiring separate training for each object. This comparison focuses on the core generation architecture, which is a standard and fair evaluation method in the field because it isolates the actual overhead of the model in real-time interactive loops. Therefore, this comparison fairly excludes the individual upstream, one-off costs of all models (including ours), such as generic part segmentation preprocessing. Similarly, the overhead of (one-time) VLM calls used for semantic reasoning in our framework was intentionally moved out of this real-time loop by our decoupling design, thus boosting efficiency.

\section{Conclusion}
In this paper, we propose DragMesh, a lightweight framework designed to bridge the gap between static 3D generation and real-time physics interaction. Current methods are either too slow or lack physical consistency.
We address this dilemma with a decoupled framework that separates semantic reasoning from geometric generation. We first leverage an LLM for semantic reasoning to determine the joint type. This intent is then fed into our kinematic prediction network (KPP-Net), which performs geometric reasoning to regress the precise joint axes and origin.
Finally, a novel, lightweight core built around this—the FiLM-conditional Dual Quaternion VAE—generates physically consistent and temporally smooth animation conditioned on these complete kinematic parameters.

\clearpage
\clearpage
{
    \small
    \bibliographystyle{ieeenat_fullname}
    \bibliography{main}

@String(CVPR= {IEEE Conf. Comput. Vis. Pattern Recog.})

@String(ICCV= {Int. Conf. Comput. Vis.})

@String(ECCV= {Eur. Conf. Comput. Vis.})

@String(ICLR = {Int. Conf. Learn. Represent.})

@String(CVPR  = {CVPR})

@String(ICCV  = {ICCV})

@String(ECCV  = {ECCV})

@String(ICLR  = {ICLR})

@InProceedings{Park_2019_CVPR,
author = {Park, Jeong Joon and Florence, Peter and Straub, Julian and Newcombe, Richard and Lovegrove, Steven},
title = {DeepSDF: Learning Continuous Signed Distance Functions for Shape Representation},
booktitle = CVPR,
year = {2019}
}

@inproceedings{pan2023_DragGAN,
    title={Drag Your GAN: Interactive Point-based Manipulation on the Generative Image Manifold}, 
    author={Pan, Xingang and Tewari, Ayush and Leimk{\"u}hler, Thomas and Liu, Lingjie and Meka, Abhimitra and Theobalt, Christian},
    booktitle = {ACM SIGGRAPH},
    year={2023}
}

@inproceedings{shi2024dragdiffusion,
  title={Dragdiffusion: Harnessing diffusion models for interactive point-based image editing},
  author={Shi, Yujun and Xue, Chuhui and Liew, Jun Hao and Pan, Jiachun and Yan, Hanshu and Zhang, Wenqing and Tan, Vincent YF and Bai, Song},
  booktitle = CVPR,
  pages={8839--8849},
  year={2024}
}

@inproceedings{wu2024draganything,
  title={Draganything: Motion control for anything using entity representation},
  author={Wu, Weijia and Li, Zhuang and Gu, Yuchao and Zhao, Rui and He, Yefei and Zhang, David Junhao and Shou, Mike Zheng and Li, Yan and Gao, Tingting and Zhang, Di},
  booktitle = ECCV,
  pages={331--348},
  year={2024}
}

@inproceedings{li2024dragapart,
  title={Dragapart: Learning a part-level motion prior for articulated objects},
  author={Li, Ruining and Zheng, Chuanxia and Rupprecht, Christian and Vedaldi, Andrea},
  booktitle=ECCV,
  pages={165--183},
  year={2024}
}

@inproceedings{liu2025building,
  title={Building Interactable Replicas of Complex Articulated Objects via Gaussian Splatting},
  author={Liu, Yu and Jia, Baoxiong and Lu, Ruijie and Ni, Junfeng and Zhu, Song-Chun and Huang, Siyuan},
  booktitle=ICLR,
  year={2025},
}

@InProceedings{chen2025freeart3d,
  title = {FreeArt3D: Training-Free Articulated Object Generation using 3D Diffusion},
  author = {Chen, Chuhao and Liu, Isabella and Wei, Xinyue and Su, Hao and Liu, Minghua},
  booktitle = {SIGGRAPH Asia 2025 Conference Papers},
  year = {2025}
}

@inproceedings{su2025artformer,
  title={Artformer: Controllable generation of diverse 3d articulated objects},
  author={Su, Jiayi and Feng, Youhe and Li, Zheng and Song, Jinhua and He, Yangfan and Ren, Botao and Xu, Botian},
  booktitle=CVPR,
  pages={1894--1904},
  year={2025}
}

@inproceedings{fan2017point,
  title={A point set generation network for 3d object reconstruction from a single image},
  author={Fan, Haoqiang and Su, Hao and Guibas, Leonidas J},
  booktitle=CVPR,
  pages={605--613},
  year={2017}
}

@inproceedings{gao2025meshart,
    title = {MeshArt: Generating Articulated Meshes with Structure-guided Transformers},
    author = {Gao, Daoyi and Siddiqui, Yawar and Li, Lei and Dai, Angela},
    booktitle = CVPR,
    year = {2025}
}

@inproceedings{gao2025partrm,
  title={PartRM: Modeling Part-Level Dynamics with Large Cross-State Reconstruction Model},
  author={Gao, Mingju and Pan, Yike and Gao, Huan-ang and Zhang, Zongzheng and Li, Wenyi and Dong, Hao and Tang, Hao and Yi, Li and Zhao, Hao},
  booktitle = CVPR,
  pages={7004--7014},
  year={2025}
}

@article{chen2024mvdrag3d,
  title={Mvdrag3d: Drag-based creative 3d editing via multi-view generation-reconstruction priors},
  author={Chen, Honghua and Lan, Yushi and Chen, Yongwei and Zhou, Yifan and Pan, Xingang},
  journal={arXiv preprint arXiv:2410.16272},
  year={2024}
}

@inproceedings{zhang2023adding,
  title={Adding conditional control to text-to-image diffusion models},
  author={Zhang, Lvmin and Rao, Anyi and Agrawala, Maneesh},
  booktitle=CVPR,
  pages={3836--3847},
  year={2023}
}

@article{peng2024controlnext,
  title={Controlnext: Powerful and efficient control for image and video generation},
  author={Peng, Bohao and Wang, Jian and Zhang, Yuechen and Li, Wenbo and Yang, Ming-Chang and Jia, Jiaya},
  journal={arXiv preprint arXiv:2408.06070},
  year={2024}
}

@article{qiu2025articulate,
    title={Articulate AnyMesh: Open-vocabulary 3D Articulated Objects Modeling}, 
    author={Qiu, Xiaowen and Yang, Jincheng and Wang, Yian and Chen, Zhehuan and Wang, Yufei and Wang, Tsun-Hsuan and Xian, Zhou and Gan, Chuang},
    journal={arXiv preprint arXiv:2502.02590}, 
    year={2025} 
}

@inproceedings{siddiqui2024meshgpt,
  title={Meshgpt: Generating triangle meshes with decoder-only transformers},
  author={Siddiqui, Yawar and Alliegro, Antonio and Artemov, Alexey and Tommasi, Tatiana and Sirigatti, Daniele and Rosov, Vladislav and Dai, Angela and Nie{\ss}ner, Matthias},
  booktitle= CVPR,
  pages={19615--19625},
  year={2024}
}

@article{liu2025artgs,
  title={Artgs: Building interactable replicas of complex articulated objects via gaussian splatting},
  author={Liu, Yu and Jia, Baoxiong and Lu, Ruijie and Ni, Junfeng and Zhu, Song-Chun and Huang, Siyuan},
  journal={arXiv preprint arXiv:2502.19459},
  year={2025}
}

@article{chen2024meshanything,
  title={Meshanything: Artist-created mesh generation with autoregressive transformers},
  author={Chen, Yiwen and He, Tong and Huang, Di and Ye, Weicai and Chen, Sijin and Tang, Jiaxiang and Chen, Xin and Cai, Zhongang and Yang, Lei and Yu, Gang and others},
  journal={arXiv preprint arXiv:2406.10163},
  year={2024}
}

@article{chen2024meshanything2,
  title={Meshanything v2: Artist-created mesh generation with adjacent mesh tokenization},
  author={Chen, Yiwen and Wang, Yikai and Luo, Yihao and Wang, Zhengyi and Chen, Zilong and Zhu, Jun and Zhang, Chi and Lin, Guosheng},
  journal={arXiv preprint arXiv:2408.02555},
  year={2024}
}

@article{chen2024partgen,
  title={PartGen: Part-level 3D Generation and Reconstruction with Multi-View Diffusion Models},
  author={Minghao Chen and Roman Shapovalov and Iro Laina and Tom Monnier and Jianyuan Wang and David Novotny and Andrea Vedaldi},
  journal={arXiv preprint arXiv:2412.18608},
  year={2024}
}

@article{hong2023lrm,
  title={Lrm: Large reconstruction model for single image to 3d},
  author={Hong, Yicong and Zhang, Kai and Gu, Jiuxiang and Bi, Sai and Zhou, Yang and Liu, Difan and Liu, Feng and Sunkavalli, Kalyan and Bui, Trung and Tan, Hao},
  journal={arXiv preprint arXiv:2311.04400},
  year={2023}
}

@article{tochilkin2024triposr,
  title={Triposr: Fast 3d object reconstruction from a single image},
  author={Tochilkin, Dmitry and Pankratz, David and Liu, Zexiang and Huang, Zixuan and Letts, Adam and Li, Yangguang and Liang, Ding and Laforte, Christian and Jampani, Varun and Cao, Yan-Pei},
  journal={arXiv preprint arXiv:2403.02151},
  year={2024}
}

@article{li2025triposg,
  title={TripoSG: High-Fidelity 3D Shape Synthesis using Large-Scale Rectified Flow Models},
  author={Li, Yangguang and Zou, Zi-Xin and Liu, Zexiang and Wang, Dehu and Liang, Yuan and Yu, Zhipeng and Liu, Xingchao and Guo, Yuan-Chen and Liang, Ding and Ouyang, Wanli and others},
  journal={arXiv preprint arXiv:2502.06608},
  year={2025}
}

@inproceedings{mildenhall2020nerf,
 title={NeRF: Representing Scenes as Neural Radiance Fields for View Synthesis},
 author={Ben Mildenhall and Pratul P. Srinivasan and Matthew Tancik and Jonathan T. Barron and Ravi Ramamoorthi and Ren Ng},
 year={2020},
 booktitle=ECCV,
}

@InProceedings{Lionar_2025_CVPR,
    author    = {Lionar, Stefan and Liang, Jiabin and Lee, Gim Hee},
    title     = {TreeMeshGPT: Artistic Mesh Generation with Autoregressive Tree Sequencing},
    booktitle = CVPR,
    year      = {2025},
    pages     = {26608-26617}
}

@InProceedings{Peng_2022_CVPR,
    author    = {Peng, Kebin and Islam, Rifatul and Quarles, John and Desai, Kevin},
    title     = {TMVNet: Using Transformers for Multi-View Voxel-Based 3D Reconstruction},
    booktitle = {CVPR Workshops},
    year      = {2022}
}

@inproceedings{Zhou_2021_ICCV,
    author    = {Zhou, Linqi and Du, Yilun and Wu, Jiajun},
    title     = {3D Shape Generation and Completion Through Point-Voxel Diffusion},
    booktitle = ICCV,
    year      = {2021},
    pages     = {5826-5835}
}

@inproceedings{zhao2025deepmesh,
  title={DeepMesh: Auto-Regressive Artist-mesh Creation with Reinforcement Learning},
  author={Zhao, Ruowen and Ye, Junliang and Wang, Zhengyi and Liu, Guangce and Chen, Yiwen and Wang, Yikai and Zhu, Jun},
  booktitle = ICCV,
  year={2025}
}

@inproceedings{chen2023shaddr,
  title={ShaDDR: interactive example-based geometry and texture generation via 3D shape detailization and differentiable rendering},
  author={Chen, Qimin and Chen, Zhiqin and Zhou, Hang and Zhang, Hao},
  booktitle={SIGGRAPH Asia 2023 Conference Papers},
  pages={1--11},
  year={2023}
}

@article{li2024craftsman3d,
  title={Craftsman3d: High-fidelity mesh generation with 3d native generation and interactive geometry refiner},
  author={Li, Weiyu and Liu, Jiarui and Yan, Hongyu and Chen, Rui and Liang, Yixun and Chen, Xuelin and Tan, Ping and Long, Xiaoxiao},
  journal={arXiv preprint arXiv:2405.14979},
  year={2024}
}

@article{xu2024instantmesh,
  title={Instantmesh: Efficient 3d mesh generation from a single image with sparse-view large reconstruction models},
  author={Xu, Jiale and Cheng, Weihao and Gao, Yiming and Wang, Xintao and Gao, Shenghua and Shan, Ying},
  journal={arXiv preprint arXiv:2404.07191},
  year={2024}
}

@article{wang2024llamameshunifying3dmesh,
      title={LLaMA-Mesh: Unifying 3D Mesh Generation with Language Models}, 
      author={Zhengyi Wang and Jonathan Lorraine and Yikai Wang and Hang Su and Jun Zhu and Sanja Fidler and Xiaohui Zeng},
      year={2024},
      journal={arXiv preprint arXiv:2411.09595},
}

@article{ma2025p3,
  title={P3-sam: Native 3d part segmentation},
  author={Ma, Changfeng and Li, Yang and Yan, Xinhao and Xu, Jiachen and Yang, Yunhan and Wang, Chunshi and Zhao, Zibo and Guo, Yanwen and Chen, Zhuo and Guo, Chunchao},
  journal={arXiv preprint arXiv:2509.06784},
  year={2025}
}

@article{xie20252d,
  title={2D Instance Editing in 3D Space},
  author={Xie, Yuhuan and Pan, Aoxuan and Lin, Ming-Xian and Huang, Wei and Huang, Yi-Hua and Qi, Xiaojuan},
  journal={arXiv preprint arXiv:2507.05819},
  year={2025}
}

@inproceedings{pandey2024diffusion,
  title={Diffusion handles enabling 3d edits for diffusion models by lifting activations to 3d},
  author={Pandey, Karran and Guerrero, Paul and Gadelha, Matheus and Hold-Geoffroy, Yannick and Singh, Karan and Mitra, Niloy J},
  booktitle=CVPR,
  pages={7695--7704},
  year={2024}
}

@article{wu2025dipo,
  title={Dipo: Dual-state images controlled articulated object generation powered by diverse data},
  author={Wu, Ruiqi and Wang, Xinjie and Liu, Liu and Guo, Chunle and Qiu, Jiaxiong and Li, Chongyi and Huang, Lichao and Su, Zhizhong and Cheng, Ming-Ming},
  journal={arXiv preprint arXiv:2505.20460},
  year={2025}
}

@article{yu2025part,
  title={Part$^{2}$GS: Part-aware Modeling of Articulated Objects using 3D Gaussian Splatting},
  author={Yu, Tianjiao and Shah, Vedant and Wahed, Muntasir and Shen, Ying and Nguyen, Kiet A and Lourentzou, Ismini},
  journal={arXiv preprint arXiv:2506.17212},
  year={2025}
}

@article{zhao2025high,
  title={High-Fidelity Simulated Data Generation for Real-World Zero-Shot Robotic Manipulation Learning with Gaussian Splatting},
  author={Zhao, Haoyu and Zeng, Cheng and Zhuang, Linghao and Zhao, Yaxi and Xue, Shengke and Wang, Hao and Zhao, Xingyue and Li, Zhongyu and Li, Kehan and Huang, Siteng and others},
  journal={arXiv preprint arXiv:2510.10637},
  year={2025}
}

@article{geng2022gapartnet,
  title={GAPartNet: Cross-Category Domain-Generalizable Object Perception and Manipulation via Generalizable and Actionable Parts},
  author={Geng, Haoran and Xu, Helin and Zhao, Chengyang and Xu, Chao and Yi, Li and Huang, Siyuan and Wang, He},
  journal={arXiv preprint arXiv:2211.05272},
  year={2022}
}

@article{he2025sparseflex,
  title={Sparseflex: High-resolution and arbitrary-topology 3d shape modeling},
  author={He, Xianglong and Zou, Zi-Xin and Chen, Chia-Hao and Guo, Yuan-Chen and Liang, Ding and Yuan, Chun and Ouyang, Wanli and Cao, Yan-Pei and Li, Yangguang},
  journal={arXiv preprint arXiv:2503.21732},
  year={2025}
}

@article{kerbl20233d,
  title={3D Gaussian splatting for real-time radiance field rendering.},
  author={Kerbl, Bernhard and Kopanas, Georgios and Leimk{\"u}hler, Thomas and Drettakis, George},
  journal={ACM Trans. Graph.},
  pages={139--1},
  year={2023}
}

@article{liu2024meshformer,
  title={MeshFormer: High-Quality Mesh Generation with 3D-Guided Reconstruction Model},
  author={Minghua Liu and Chong Zeng and Xinyue Wei and Ruoxi Shi and Linghao Chen and Chao Xu and Mengqi Zhang and Zhaoning Wang and Xiaoshuai Zhang and Isabella Liu and Hongzhi Wu and Hao Su},
  journal={arXiv preprint arXiv:2408.10198},
  year={2024}
}

@inproceedings{guo2025articulatedgs,
  title={Articulatedgs: Self-supervised digital twin modeling of articulated objects using 3d gaussian splatting},
  author={Guo, Junfu and Xin, Yu and Liu, Gaoyi and Xu, Kai and Liu, Ligang and Hu, Ruizhen},
  booktitle=CVPR,
  pages={27144--27153},
  year={2025}
}

@article{wu2025reartgs,
  title={Reartgs: Reconstructing and generating articulated objects via 3d gaussian splatting with geometric and motion constraints},
  author={Wu, Di and Liu, Liu and Linli, Zhou and Huang, Anran and Song, Liangtu and Yu, Qiaojun and Wu, Qi and Lu, Cewu},
  journal={arXiv preprint arXiv:2503.06677},
  year={2025}
}

@InProceedings{Xiang_2020_SAPIEN,
author = {Xiang, Fanbo and Qin, Yuzhe and Mo, Kaichun and Xia, Yikuan and Zhu, Hao and Liu, Fangchen and Liu, Minghua and Jiang, Hanxiao and Yuan, Yifu and Wang, He and Yi, Li and Chang, Angel X. and Guibas, Leonidas J. and Su, Hao},
title = {{SAPIEN}: A SimulAted Part-based Interactive ENvironment},
booktitle = CVPR,
year = {2020}
}

@InProceedings{Mo_2019_CVPR,
    author = {Mo, Kaichun and Zhu, Shilin and Chang, Angel X. and Yi, Li and Tripathi, Subarna and Guibas, Leonidas J. and Su, Hao},
    title = {{PartNet}: A Large-Scale Benchmark for Fine-Grained and Hierarchical Part-Level {3D} Object Understanding},
    booktitle = CVPR,
    year = {2019}
}

@misc{liu2022akb48realworldarticulatedobject,
      title={AKB-48: A Real-World Articulated Object Knowledge Base}, 
      author={Liu Liu and Wenqiang Xu and Haoyuan Fu and Sucheng Qian and Yang Han and Cewu Lu},
      year={2022},
      archivePrefix={arXiv},
}
}

\clearpage
\setcounter{page}{1}
\maketitlesupplementary

\section{Implementation Details}

\label{Comparative Study Detail}

For our comparative study \ref{Comparative Study}, we evaluated DragMesh against several baselines, including MeshArt, PartRM. We used the official, publicly available code for all methods.

To enable a fair qualitative comparison, we performed custom data preparation work tailored to each baseline method. This custom processing was necessary to adapt our data to meet the specific input requirements of each model, enabling them to achieve a minimal level of interactive functionality on our dataset. A critical step in constructing our interaction dataset was generating corresponding drag points and drag vectors for each articulated part. Our generation pipeline is as follows:

First, we parsed the kinematic data from our curated set of interactive models and converted it into a JSON format. Subsequently, we randomly sampled points $P_{start}$ on the surface of each movable part to serve as the drag points.

To generate the corresponding drag vector $V_{drag}$, we calculated the point's motion using the trajectory defined in the JSON file. We identified a key challenge: for large-angle rotations (e.g., exceeding 180°), defining $V_{drag}$ by simply subtracting the start point from the end point of the full trajectory introduces ambiguity and fails to yield a unique, correct direction.

To ensure the rigor of our data and the absolute correctness of the motion, we instead adopted a strategy based on instantaneous motion. We strictly define the "end point" $P_{end}$ as the position of the start point $P_{start}$ at the immediately subsequent timestep ($t+1$). Thus, the drag vector is $V_{drag} = P_{end} - P_{start}$, which guarantees that $V_{drag}$ accurately reflects the point's local direction of motion.

Finally, we recognized that the magnitude of this instantaneous vector might be too short to simulate a complete user interaction. We employed two strategies to generate the final vector: (1) accumulating the displacement over several subsequent timesteps, contingent on the point's trajectory remaining relatively linear (i.e., without significant deviation), or (2) directly applying a fixed scaling factor to the original instantaneous vector $V_{drag}$ to extend it to a standard length.

\paragraph{PartRm}

The data requirements of the PartRM baseline model are significantly incompatible with our workflow. The model relies on a set of discrete 3D states (e.g., state 0 to state 5) as input, while our Objaverse corpus consists of GLB models and continuous Trajectory JSON trajectories. Therefore, generating effective qualitative comparisons for PartRM requires a customized data adaptation pipeline and modifications to its core code.

We implemented a data generation pipeline to bridge PartRM’s discrete state requirements with our continuous trajectory data: First, we take state 0 (the stationary state) as input, parse the Trajectory JSON, and automatically render all six discrete states (state 0 to state 5) required by PartRM, each containing 12 views. Next, we implemented the logic for automatically solving 2D drag vectors from the 3D trajectory data and generated the propagated drags index file necessary for the PartRM evaluation process.

In addition, we modified its data loader to support the use of our newly generated propagated drags as interactive guides when a mesh is missing. When a Zero123++ image is detected as missing, the loader automatically falls back to using the original Blender-rendered image, allowing the process to continue. It is worth noting that due to the significant differences between PartRM's input requirements and our data source, even after adaptation, various issues such as missing 2D drag vectors due to insufficient information still occur.

\paragraph{ArtGS}

In our data preprocessing pipeline, we utilize Blender scripts to render multi-view images and their corresponding depth maps for each model in the dataset. During initial experiments, we observed that some default camera poses were positioned too low or too close, resulting in heavily blurred or partially occluded images that failed to capture effective geometric information.

To address this, we implemented a targeted adjustment to our rendering strategy, moving these specific camera perspectives to a farther position. However, this adjustment introduced a downstream challenge: during the 3DGS backpropagation process, these distanced views would sometimes encounter a "no visible Gaussians" issue due to sparse points, leading to abnormal gradient calculations.

To mitigate this, we adopted a trade-off training strategy: during training iterations, when the model detects these views that cannot be effectively backpropagated due to camera positioning, the system automatically skips the corresponding depth loss and uses only the RGB loss for photometric supervision. While this adaptive loss mechanism ensured effective gradient flow, allowing the model to complete training and avoid a pipeline crash, we must also note that the quality of the views generated using only this RGB supervision was exceptionally poor, with a significant loss of geometric detail.

\begin{figure*}[t]
    \centering
    \includegraphics[width=\linewidth]{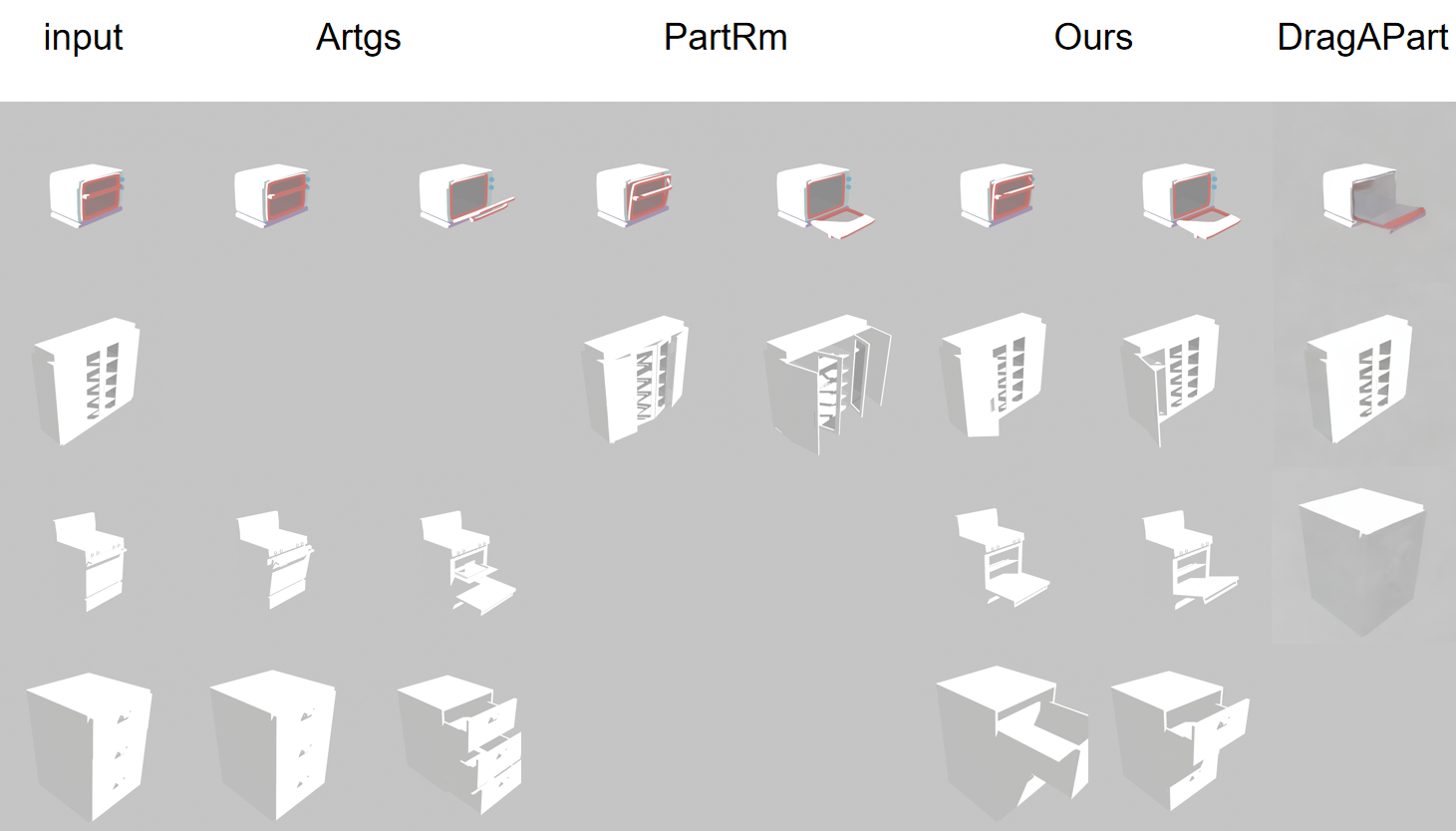} 
    \caption{Qualitative Comparison. Our method (Ours) generates plausible interactions across all categories. Blank spaces for baselines (e.g., ArtGS, PartRm, DragApart) represent generation failures, where results are omitted due to unrecognizable outlines.}
    \label{Visualization Results} 
\end{figure*}

\paragraph{DragAPart}

The DragAPart baseline model operates primarily in 2D image space, making its data preparation pipeline distinct. We first rendered the 3D GLB files from Objaverse into a static 2D RGB image, which serves as the model's conditional input.

To generate the required 2D drags tensor, we initially attempted to project our 3D kinematic data. We parsed our Trajectory JSON files, calculated a 3D motion vector, and then normalized this 3D information for the 2D plane. However, we found that for a model with no 3D perception, this 3D-to-2D normalization process introduced significant information deviation and larger errors.

Consequently, we abandoned this approach and adopted a direct 2D-native strategy. We chose to pick points directly on the 2D object (the RGB image), selecting start and end points based on our own intended interaction. This manually-defined 2D vector was then fed into the model as the drags tensor to process the single RGB image.

\paragraph{Ours}
During the training data preparation phase, we made two key modifications to the data to meet the specific architectural requirements of the DragMesh model. After acquiring the drag points and trajectory information, we first forced each object to perform single-joint interactions, as our model can only handle the movement of one joint at a time. We bypassed the VLM used for motion type classification. During training, we employed a heuristic annotation strategy: we forced the application of translation and rotation strategies to each model, and then intuitively selected and manually labeled the object with a unique, correct, and reasonable interaction type.

\section{Visualization Results}

In our qualitative comparison~\ref{Visualization Results}, we must also note that certain baseline methods fail completely when processing specific interactive data. For these cases, we mark the result as blank, as the generated effect was extremely poor (e.g., the outline information was entirely unrecognizable) and provided no basis for a meaningful comparison.

\section{Limitations and Future Work}
\subsection{Limitations}
\label{sec:limitations}
While the DragMesh framework achieves real-time, generalizable articulation, our approach currently presents several limitations which guide future research.

\begin{enumerate}
    \item The model's kinematic scope is restricted to single-joint interactions involving only two fundamental motion types: simple translation and rotation around a fixed axis, lacking the capacity for more complex features such as screw motion or multi-joint chains.
    \item Achieving robust kinematic prediction is highly sensitive to the initial geometric input. Our pipeline requires strict adherence to the expected data representation; otherwise, errors in prerequisite information, such as incorrect joint axis prediction, can arise, resulting in physically implausible mesh transformations.
    \item Our annotation-free inference pipeline’s ability to classify the motion type (e.g., revolute vs. prismatic) currently relies on the assistance of external Vision-Language Models (VLMs) for semantic reasoning. While integrating VLMs represents the current mainstream approach for handling semantic intent, we acknowledge that their inherent error rate and external dependency impact the robustness of our prediction pipeline. Removing this reliance to create a fully self-contained, robust geometric model remains a core challenge for future work.
\end{enumerate}

\subsection{Future work}

Interactive generation is progressively gaining traction in the field. In essence, it can be viewed as a key component of a World Model, that is, a system processing the behavior of a scene at the next timestep, given a specific condition. We believe a core trend for future world models, regardless of the scenario, will be a primary focus on lightweight generation. This trend is particularly critical for realizing digital Human-Object Interaction (HOI) and for robotics applications in the sim2real domain. These frontier domains not only demand a deeper understanding of physical laws (such as kinematics and dynamics) but also impose extremely low-latency requirements on response speed.

Therefore, future work must focus on finding a new equilibrium between the depth of physical understanding and the speed of lightweight generation.

For DragMesh, this specifically means:

\begin{enumerate}
    \item  VLM Dependency: Our current inference pipeline relies on an external VLM for semantic intent classification. A primary future task is to explore methods for "internalizing" this reasoning capability, perhaps through more powerful geometric encoders or by leveraging end-to-end interaction data, to achieve a truly lightweight and self-contained framework.
    \item Expanding Kinematic Scope: The model is currently limited to single-joint translation and rotation. The next step is to extend this to more complex motions, such as screw motion and multi-joint kinematic chains.
    \item  Enhancing System Robustness: Future work must address the gap between training (which uses ground-truth data) and inference (which relies on upstream predictions ). This involves making the model less sensitive to potential errors from upstream part segmentation or noisy joint parameter predictions.
\end{enumerate}

The goal is to explore how to imbue models with robust physical common sense while maintaining a compact and efficient architecture, enabling truly real-time, physically plausible intelligent agents.

\end{document}